\documentclass{article}
\usepackage{ijcai21}

\usepackage{times}
\usepackage{soul}
\usepackage{url}
\usepackage[hidelinks]{hyperref}
\usepackage[utf8]{inputenc}
\usepackage[small]{caption}
\usepackage{graphicx}
\usepackage{amsmath}
\usepackage{amsthm}
\usepackage{booktabs}
\usepackage{amssymb}
\usepackage[ruled,vlined,linesnumbered]{algorithm2e}
\usepackage{comment}

\title{Compilation-based Solvers for Multi-Agent Path Finding:\\a Survey, Discussion, and Future Opportunities}

\author{
    Pavel Surynek
    \affiliations
    Faculty of Information Technology\\     
    Czech Technical University in Prague\\
    Th\'{a}kurova 9, 160 00 Praha 6, Czechia\\    
    \emails
    pavel.surynek@fit.cvut.cz
}

\begin{document}

\maketitle

\begin{abstract}
Multi-agent path finding (MAPF) attracts considerable attention in artificial intelligence community as well as in robotics, and other fields such as warehouse logistics. The task in the standard MAPF is to find paths through which agents can navigate from their starting positions to specified individual goal positions. The combination of two additional requirements makes the problem computationally challenging: (i) agents must not collide with each other and (ii) the paths must be optimal with respect to some objective.
  
Two major approaches to optimal MAPF solving include (1) dedicated search-based methods, which solve MAPF directly, and (2) compilation-based methods that reduce a MAPF instance to  an instance in a different well established formalism, for which an efficient solver exists. The compilation-based MAPF solving can benefit from advancements accumulated during the  development of the target solver often decades long.
 
We summarize and compare contemporary compilation-based solvers for MAPF using formalisms like ASP, MIP, and SAT. We show the lessons learned from past developments and current trends in the topic and discuss its wider impact.
\end{abstract}

\section{Introduction}

Compilation is one of the most prominent techniques used across variety of computing fields ranging from theory to practice. In the context of problem solving in artificial intelligence, compilation is represented by reduction of an input instance from its source formalism to a different usually well established formalism for which an efficient solver exists. The reduction between the formalisms is usually fast so that obtaining the instance in the target formalism and interpreting the solution back to the source formalism after using the solver consumes only small part of the total runtime while the most of time is assumed to be spent by the solver. The key idea behind using compilation in problem solving is that the solving process benefits from the advancements in the solver for the target formalism, often accumulated over decades.

The compilation-based solving approach has been applied successfully applied in solving combinatorial problems like {\em plannnig} \cite{DBLP:books/daglib/0014222}, {\em verification} \cite{DBLP:books/daglib/0019162}, or {\em scheduling} \cite{DBLP:reference/crc/BlazewiczBF04} where the target formalism is often represented by {\em Boolean satisfiability} (SAT) \cite{DBLP:series/faia/BarrettSST09}, {\em mixed integer linear programming} (MIP) \cite{DBLP:books/daglib/0023873,rader2010deterministic}, {\em answer set programming} (ASP) \cite{DBLP:books/sp/Lifschitz19} or {\em constraint satisfaction} (CSP) \cite{DBLP:books/daglib/0016622}. Significant advancements has been achieved in compilation-based approaches in specific domains, namely in {\em multi-agent path finding} (MAPF) \cite{DBLP:conf/aiide/Silver05,DBLP:journals/jair/Ryan08,DBLP:conf/aaai/Standley10} that we are focusing on in this paper.

The standard variant of MAPF is the problem of finding collision-free paths for a set of agents from their starting positions to individual goal positions. Agents move in an environment which is usually modeled as an undirected graph $G=(V,E)$, where vertices represent positions and edges the possibility of moving between positions. Agents in this abstraction are discrete items, commonly denoted $A=\{a_1,a_2,...,a_k\}, k \leq |V|$, placed in vertices of the graph, moving instantaneously between vertices provided that there is always at most one agent in a vertex and no two agents traverse an edge in opposite directions \footnote{Different movement rules exist such permitting the move into a vacant vertex only. There is also large body of works dealing with related problems like {\em pebble motion} in graphs \cite{DBLP:conf/focs/KornhauserMS84}, {\em token swapping} and {\em token permutations} in graphs \cite{DBLP:journals/algorithmica/BonnetMR18} using similar movement primitives.}.

There are numerous applications of the MAPF problem in warehouse logistics \cite{DBLP:conf/atal/LiTKDKK20}, traffic optimization \cite{mohanty2020flatlandrl}, multi-robot systems \cite{DBLP:conf/icra/PreissHSA17}, and computer games \cite{DBLP:conf/aaai/SnapeGBLM12} to name few representatives, for more real-life applications see \cite{DBLP:conf/socs/FelnerSSBGSSWS17}.

Relatively simple formulation of the MAPF problem is an important factor that made it an accessible target of various solving methods including compilation-based approaches. The simultaneous existence of diverse methods for MAPF has fostered mutual cross-fertilization and deeper understanding. Current state-of-the-art compilation-based solvers for MAPF go even beyond the standard single shot {\bf reduction-solving-interpreting} loop coined for classical planning by the SATPlan algorithm \cite{DBLP:conf/ecai/KautzS92} where Boolean satisfiability has been used as the target formalism and the SAT solver has been treated merely as a black-box solver. Intensive cross-fertilization between the dedicated search-based methods for MAPF, that solve the problem directly, and compilation techniques led to numerous improvements in encodings of the problem in target formalisms but also how the solver is used, treating it less like a black-box.

The effort culminated recently in a combination of compilation and lazy conflict resolution introduced in Conflict-based search algorithm (CBS) \cite{DBLP:journals/ai/SharonSFS15} resulting in approaches that construct the target encoding lazily in close cooperation with the solver. The solver in these lazy schemes suggests solutions for incomplete encodings of the input instance that do not specify it fully. After checking the interpreted solution against the original specification, that is if it is a valid MAPF solution (agents do not collide, do not jump, do not disappear, and do not appear spontaneously, etc.), the high-level part of the MAPF solver suggests a refinement of the encoding and the process is repeated. This scheme has been implemented using SAT \cite{DBLP:conf/ijcai/Surynek19} and MIP \cite{DBLP:conf/aips/GangeHS19,DBLP:conf/ijcai/LamBHS19} as target formalisms.

We briefly summarize in this paper how the research in compilation for MAPF came to this point and discuss in more details opportunities and limitations of contemporary best SAT-based and MIP-based compilation schemes.

\section{An Overview of Target Formalisms}

We will first give a brief overview of popular formalisms that are often used as the target of the compilation process.

\noindent {\bf Constraint satisfaction problem (CSP)} is a tuple $(X,D,C)$ where $X$ is a finite set of variables, $D$ is a finite domain of values that can be assigned to variables, and $C$ is a set of constraints in the form of arbitrary relations over the variables, that is a constraint is a subset of Cartesian product $D \times D \times ... D$ (the arity corresponds to how many variables participate in the constraint). A solution of CSP is an assignment of values from $D$ to all variables such that all constraints are satisfied by this assignment (that is, constraints are treated as a {\bf conjunction}).

In the MAPF for example, we can express positions of agents $a_i$ at time step $t$ using a variable $X_i^t \in V$ (indicating that the domain is $V$) \cite{DBLP:conf/icra/Ryan10}. Then constraints expressing the MAPF rules such as: $X_i^t \neq X_i^{t+1} \rightarrow \{ X_i^t, X_i^{t+1} \} \in E$, agents use edges (do not skip); and $\mathit{Distinct}(X_1^t,X_2^t, ..., X_k^t)$, agents do not collide in vertices, etc.. In addition to MAPF rules, constraints bounding the objectives are added to obtain encoding of the bounded MAPF.

The CSP paradigm provides powerful complete search algorithms coupled with constraint propagation and domain filtering techniques that prune the search space \cite{DBLP:conf/cp/JussienDB00}. One of the most significant advantages of CSP in contrast to other paradigms are {\em global constraints}, dedicated filtering algorithms often based on matchings for special relations like $\mathit{Distinct}$ (variables should take different values) or more general cardinality constraints, that enforce consistency across large set of variables \cite{DBLP:conf/aaai/Regin94}. Using global constraints it is easy to detect that $\mathit{Distinct}(X_1^t,X_2^t, X_3^t)$ cannot be satisfied when $D = \{v_1, v_2\}$ (there is no matching between the variables and values of size 3) but it is hard to see it looking on the individual inequalities $X_i^t \neq X_j^t$ separately.

\noindent {\bf Mixed integer linear programming (MIP)}. A linear program (LP) is a finite list of linear inequalities plus linear objective, the task is to minimize the objective such that the inequalities hold. Geometrically the inequalities define a polytope and the objective function defines a gradient so the task is to find a some boundary point of the polytope that minimizes the gradient. Formally, the task is to minimize $c^\top x$ subject to $Px \leq b$, $x \geq 0$, where $x$ is a real vector representing the decision variables, $P$ is a matrix of real coefficients of linear inequalities, $b$ is a another real-valued vector. An optimal solution to linear program can be found in polynomial time however in discrete decision problems like MAPF it not much convenient to use fractional assignment of decision variables.

That is why often some or all decision variables are declared to be integers making LP an integer program (IP) or mixed integer program (MIP) respectively. Introducing integer decision variables improves the expressive power for discrete problems but on the contrary in makes the problem NP-hard.

The important advantage of MIP is natively supported arithmetic hence optimizing with respect to various cumulative objectives used in MAPF can be modeled easily when MIP is used as the target formalism.

\noindent {\bf Boolean satisfiability (SAT)} problem consists in deciding whether there exists truth-value assignment of variables that satisfies a given Boolean formula. The formula is often specified using the {\em conjunctive normal form} (CNF), it is a conjunction of clauses where each clause is a disjunction of literals, a literal is either a variable or a negation of a variable. SAT is often considered to be a canonical NP-complete problem \cite{DBLP:conf/stoc/Cook71}. A formula in CNF can be regarded through the CSP formalism so that clauses represent individual constraints. Such view actually catalyses mutual cross fertilization between solving algorithms for SAT and CSP. Modern SAT solvers are based on the {\em conflict-driven clause learning} algorithm \cite{DBLP:conf/sat/EenS03,DBLP:journals/ijait/AudemardS18} that actually implement constraint propagation and back-jumping techniques known from CSP.

The significant challenge when SAT is used as the target formalism is bridging the original representation of the problem and the yes/no environment of SAT. In the context of MAPF, it is difficult to handle objectives that use arithmetic.

The disadvantage in the lack of expressiveness caused by the fact that formulae are often far from being human readable, is well balanced by efficiency of SAT solvers enabled by numerous techniques such as learning, restarts, etc.

\noindent {\bf Answer set programming (ASP)}. The difficulty of encoding problems in the SAT formalism at the low-level led to development of various high-level languages that enable encoding of problems more intuitively. One such notable higher level approach used in MAPF is ASP \cite{DBLP:conf/aaai/ErdemKOS13}. The problem in ASP is described as a logic program which is automatically translated to a Boolean formula. The formula is solved by the SAT solver and the outcome is propagated back at the logic program level where it is already in the human readable form.

We show a part of the ASP program for MAPF to illustrate how the ASP paradigm works, see \cite{DBLP:conf/aaai/ErdemKOS13} for the full program. We have atoms $\mathit{path}(i,t,v)$ whose interpretation is that agent $a_i \in A$ is at vertex $v$ at time step $t$. These atoms are defined recursively using the rules of a {\em logic program}:

\begin{equation}
\mathit{path}(i,0,v) \leftarrow \mathit{start}(i,v)
\label{asp-eq-1}
\end{equation}

\begin{equation}
\begin{split}
1\{\mathit{path}(i,t+1,u), \mathit{path}(i,t+1,v): \mathit{edge}(u,v)\}1 \\
\leftarrow \mathit{path}(i,t,v)
\label{asp-eq-2}
\end{split}
\end{equation}

The program says that agents' paths start at starting positions \ref{asp-eq-1} and then either traverse edge or wait in a vertex \ref{asp-eq-2}. Ones around the rule specify the number of atoms selected for the answer of the rule, the lower and the upper bounds are specified. Here one of the two instances of $\mathit{path}$ is selected.

ASP allows to specify constraints which are the rules without the head, eliminating answers that satisfy the body. Constraints can be used to eliminate conflicts for MAPF as follows:

\begin{equation}
\leftarrow \mathit{path}(i,t,v),\mathit{path}(j,t,v)
\label{asp-eq-3}
\end{equation}

The ASP program is solved via reduction to SAT and by using the SAT solver. The important advantage of ASP in contrast to plain SAT-based approach is that ASP provides high level language to describe problems and one does not need to construct the target Boolean formula. The formula is constructed automatically by the ASP solver from the logic program.

\section{From Classical Planning to MAPF}

One of the pioneering works that used compilation for problem solving is the SATPlan algorithm reducing the classical planning problem to Boolean satisfiability \cite{DBLP:conf/ecai/KautzS92}. Classical planning is the task of finding a sequence of actions that transforms a given initial state of some abstract world to a desired goal state. States are described as finite sets of atoms. Actions can add and remove atoms in states provided their preconditions are satisfied.

SATPlan is the important milestone in non-trivial compilation schemes since there is no one-to-one correspondence between the planning instance and the target SAT instance. Instead, a bounded decision variant of the input problem is introduced in which the number of time steps is bounded. The SAT solver is consulted with sequence of yes/no questions on the bounded variant, whether there exists a plan consisting of a specified number of time steps $t_{max}$.

\subsection{Time Expansion}

Another important concept significantly used by SATPlan is a time expansion that enables representing states at individual time steps inside the target formalism. Decision variables representing atoms are indexed with time steps up to $t_{max}$; the variable indexed with $0 < t \leq t_{max}$ is set to $\mathit{TRUE}$ if and only if the corresponding atom holds at time step $t$. In addition to state variables, there are action variables for each time step $t$ indicating whether given action takes place at $t$. Constraints ensure that if an action variable at $t$ is set to $TRUE$ then precondition atoms must hold in state variables at time step $t$ and effects must hold in state variables at time step $t+1$.

Consider an action that moves agent $a$ from vertex $v_1$ to vertex $v_2$: $\mathit{precond(move)}=\{at(a,v_1)\}$, $\mathit{effect^+(move)}=\{at(a,v_2)\}$, $\mathit{effect^-(move)}=\{at(a,v_1)\}$ for which we need to introduce following constraints $\forall t \in \{1,2,...,t_{max}-1\}$:
\begin{equation}
\label{eq_SATPlan-example}
move^t \rightarrow at(a,v_1)^t \wedge at(a,v_2)^{t+1} \wedge \neg at(a,v_1)^{t+1}
\end{equation}

Such encoding can be easily constructed as well as read off in the interpretation phase. The pseudo-code of SATPlan is shown as Algorithm \ref{alg-SATPlan}.

\begin{algorithm}[t]
\begin{footnotesize}
\SetKwBlock{NRICL}{SATPlan($\mathcal{P}$)}{end} \NRICL{
    $t_{\mathit{max}} \gets 1$\\
    \While {$\mathit{TRUE}$} {
        $F \gets$ encode-SAT($\mathcal{P}$,$t_{\mathit{max}}$)\\
        $\mathit{assignment} \gets$ consult-SAT-Solver($F$)\\
        \If {$\mathit{assignment} \neq \mathit{UNSAT}$}{
            $\mathit{plan} \gets$ interpret($\mathcal{P}$,$\mathit{assignmnet}$)\\
            \Return $\mathit{plan}$
        }
        $t_{\mathit{max}} \gets t_{\mathit{max}} + 1$\\
     }
}
\caption{Framework of the SATPlan algorithm.} \label{alg-SATPlan}
\end{footnotesize}
\end{algorithm}

\subsection{Encodings of MAPF}
The early compilation-based approaches to MAPF were rooted in the CSP formalism \cite{DBLP:conf/icra/Ryan10}. Even early SAT-based approaches used CSP-like variables for expressing the positions of agents with domains consisting all possible vertices \cite{DBLP:conf/ictai/Surynek12}. These decision variables were represented as a vector consisting of the logarithmic number of bits (Boolean variables) with respect to the size of the domain {\em log-space encoding} \cite{DBLP:books/sp/Petke15}. The disadvantage of log-space encodings in the context of MAPF seems to be weak Boolean constraint propagation and the fact that reachability analysis does not prune out variables from the encoding (removing values from decision variables corresponds to forbidding certain combination of settings of bit vectors but Boolean variables representing bits remain).

This is why all following SAT-based compilations of MAPF used {\em direct encoding} where there is a single Boolean variable $\mathcal{X}_{i,v}^t$ for each agent $a_i \in A$, each vertex $v \in V$, and relevant time step $t$. Despite these decision variables are somewhat redundant they provide a good support for constraint propagation and in reachability analysis they can be completely removed from the formula.

\section{Milestones Making Compilation Competitive}

Although modern SAT solvers are powerful search tools they alone are not sufficient and the way how the MAPF instance is presented to the SAT solver is equally important. From my personal perspective there were two milestones that made compilation for MAPF a competitive alternative to search-based methods: {\bf (1)} reachability analysis and {\bf (2)} introducing laziness.

\subsection{Reachability Analysis}

The disadvantage of SAT-based compilation in planning is often the big size of the formula because variables for all ground atoms and all ground actions must be included. That is why improvement such as reachability analysis have been introduced. Reachability analysis in classical planning is connected with {\em planning graphs} \cite{DBLP:journals/ai/BlumF97}. The planning graph is a structure representing time expansion of the set of possibly valid atoms across discrete time steps. Starting with the initial state at the first time step, atoms for the next time step are generated first by considering all applicable actions in the previous time step and by adding their positive effects together with atoms from the previous time step. To mitigate the growth of the set of atoms the concept of {\em mutual exclusion} (mutex) for pairs of atoms and actions are considered. Whenever two atoms are mutex they cannot satisfy the preconditions of an action and the action is hence pruned out. Mutexes originate at pairs of conflicting actions and propagate to atoms in their effects forward in the planning graph.

Reachability analysis analogous to planning-graphs in the domain of MAPF is done via {\em multi-valued decision diagrams} (MDDs) \cite{DBLP:conf/cp/AndersenHHT07}. MDDs were surprisingly not used in compilation-based approach for the first time but they were originally applied in the Increasing-cost tree search algorithm (ICTS) \cite{DBLP:journals/ai/SharonSGF13}. The idea of MDD is to represent all paths of the specified cost.

The compilation introduced in the MDD-SAT algorithm \cite{DBLP:conf/ecai/SurynekFSB16} is analogous to the more advanced variant of SATPlan with planning graphs \cite{DBLP:conf/ijcai/KautzS99}. Both algorithms use eager compilation, that is, the resulting Boolean formula is constructed in one shot and the SAT solver is regarded as a black-box. Incremental scheme to find an optimal plan is also similar in both algorithms.

The role of planning graphs for reachability analysis is substituted by {\em multi-valued decision diagrams} (MDDs). MDD$_i$ is constructed for each agent $a_i \in A$ and contains a copy of the underlying graph $G$ for every relevant time step where nodes that are not reachable because they are to far from the starting vertex or from the goal vertex are removed (for example vertex $v$ whose distance from agent's $a_i$ goal is 5 will not be included at time step $3$ in MDD$_i$ consisting of 4 time steps in total, because only 1 step remains to travel the distance of 5). Directed edges in MDDs represent move and wait actions, that is, they interconnect nodes at consecutive time steps in MDD whose corresponding vertices are either the same or connected in $G$. A directed path in MDD$_i$ corresponds to a plan for agent $a_i$.

Decision Boolean variables  $\mathcal{X}_{i,v}^t$ and  $\mathcal{E}_{i,u,v}^t$ are introduced for each node $(v,t)$ end edge $[(u,t);(v,t+1)]$ in MDD$_i$ respectively. The variables are $\mathit{TRUE}$ iff the agent uses the corresponding vertex or edge at given time step.

To encode MAPF rules constraints are introduces over these variables. We list here few examples only, for the complete list of constraints see \cite{DBLP:conf/ecai/SurynekFSB16}. Collisions in vertices can be ruled out by the following constraint for every $v \in V$ and timestep $t$:

\begin{equation}
    {\sum_{a_i \in A \:|\:(v,t) \in \mathit{MDD}_i}{\mathcal{X}_{i,v}^t} \leq 1
    }
    \label{eq_SAT-1}
\end{equation}

There are various ways how to translate the constraint using propositional clauses. One possible way is to introduce $\neg \mathcal{X}_{i,v}^t \vee \neg \mathcal{X}_{j,v}^t$ for all possible pairs of $a_i$ and $a_j$.

Following constraints ensure that directed paths are taken in MDDs. If agent $a_i$ appears in $u \in V$ at time step $t$ then it has to leave through exactly one edge connected to $u$:

\begin{equation}
   {  \mathcal{X}_{i,u}^t \rightarrow \bigvee_{[(u,t);(v,t+1)] \in MDD_i}{\mathcal{E}_{i,u,v}^t}
   }
   \label{eq_SAT-2}
\end{equation}
\begin{equation}
   {  \sum_{(v,t+1)\:|\:[(u,t);(v,t+1)] \in MDD_i}{\mathcal{E}_{i,u,v}^t \leq 1}
   }
   \label{eq_SAT-3}
\end{equation}

\subsection{Introducing Laziness}

Conflict-based search (CBS) \cite{DBLP:journals/ai/SharonSFS15} is currently the most popular approach for MAPF. It is due to its simple and elegant idea which enables to implement the algorithm relatively easily, its good performance, and openness to various improvements via using heuristics.

From the compilation perspective, the CBS algorithm should be understood as a lazy method that tries to solve an underspecified problem and relies on to be lucky to find a correct solution even using this incomplete specification. There is another mechanism that ensures soundness of this lazy approach, the branching scheme. If the CBS algorithm is not lucky, that is, the candidate solution is incorrect in terms of MAPF rules, then the search branches for each possible refinement of discovered MAPF rule violation and the refinement is added to the problem specification in each branch. Concretely, the MAPF rule violations are conflicts of pairs of agents such as collision of $a_i \in A$ and $a_j \in A$ in $v$ at time step $t$ and the refinements are conflict avoidance constraints for single agents in the form that $a_i \in A$ should avoid $v$ at time step $t$ (for $a_j$ analogously).

While in CBS, the branching scheme and the refinements must be explicitly implemented, in the compilation-based approach we can just enforce MAPF rule violation by adding a new constraint into the problem specification and leave branching to the solver for the target formalism. In the SAT case, conflict can be avoided by adding following disjunctive constraints representing both branches in CBS:

\begin{equation}
\neg \mathcal{X}_v^t(a_i) \vee \neg \mathcal{X}_v^t(a_j)
\end{equation}

In this way, the encoding of MAPF (or any other problem) is built dynamically and eventually may end up by the complete specification of the problem as done in MDD-SAT (or in SATPlan in the context of classical planning).

This approach is commonly known as {\em lazy encoding} and is often used in the context of {\em satisfiability modulo theories} (SMT) and also in ILP and MIP. In the context of SMT, we are interested in decision procedures for some complex logic theory $T$, that is decomposed into the Boolean part given to the SAT solver and decision procedure for the conjunctive fragment of $T$, denoted $\mathit{DECIDE}_T$. The SAT solver and $\mathit{DECIDE}_T$ cooperate in solving logic formula in $T$. SAT solver chooses literals, $\mathit{DECIDE}_T$ verifies and suggests refinements. In this sense, CBS is analogous to this approach where the role of $\mathit{DECIDE}_T$ is represented by conflict checking procedure.

Generally, SMT provides a view that starting constraints can be different than those defining correct paths and omitting conflicts. How do such different starting constraint perform is yet to be verified.

As shown by experiments a solution or a proof of that it does not exists is often found for incomplete specification (that is, well before all constraints are added to the encoding). Surprisingly intuitive explanation why this is possible comes from the geometry of linear programming. The finite set of inequalities define a polytope of feasible solutions as intersection of half-spaces, the boundary of the polytope are made by planes. Optimal feasible solution is an element of some of the planes defining the boundary but not an element of all of them (in other words some inequalities are satisfied because optimal solution is deep inside their half-space). Hence, to specify the optimal solution one does not need all the constraints. Similarly if there is no solution, the polytope is empty. Again this may happen by intersecting only some of the half-spaces.

The CBS inspired SAT-based compilation algorithm for MAPF known as SMT-CBS is shown as Algorithm \ref{alg-SATCompilation}.

\begin{algorithm}[t]
\begin{footnotesize}
\SetKwBlock{NRICL}{Lazy-SAT-MAPF($\mathcal{M}$)}{end} \NRICL{
    $\mathit{SoC} \gets$ lower-Bound($\mathcal{M}$)\\
    $\mathit{conflicts} \gets \emptyset$\\
    \While {$\mathit{TRUE}$} {   
        $(\mathit{paths,conflicts}) \gets$ \\ Solve-Bounded($\mathcal{M}$, $SoC$,$conflicts$)\\
        \If {$\mathit{paths} \neq \mathit{UNSAT}$}{
            \Return $\mathit{paths}$
        }
        $\mathit{SoC} \gets \mathit{SoC} + 1$\\
     }
}

\SetKwBlock{NRICL}{Solve-Bounded($\mathcal{M}$,$\mathit{SoC}$,$conflicts$)}{end} \NRICL{
	\While {$\mathit{TRUE}$} {
          $F' \gets$ encode-SAT($\mathcal{M}$,$\mathit{SoC}$,$\mathit{conflicts}$)\\        
          $\mathit{assignment} \gets$ consult-SAT-Solver($F'$)\\
          
          \If {$\mathit{assignment} \neq \mathit{UNSAT}$}{
            $\mathit{paths} \gets$ interpret($\mathcal{M}$,$\mathit{assignment}$)\\
            $\mathit{conflicts'} \gets$ validate($\mathcal{M}$,$\mathit{paths})$\\
            \If{$\mathit{conflicts'} = \emptyset$}{
                \Return $(\mathit{paths}, \mathit{conflicts})$
            }
                       
            \For{each $c \in \mathit{conflicts'}$}{
              $F' \gets F' \cup$ eliminate-Conflict($c$)\\
            }
            $\mathit{conflicts} \gets \mathit{conflicts} \cup \mathit{conflicts'}$
          }
          \Return {($\mathit{UNSAT}$,$\mathit{conflicts}$)}\\
      }
}

\caption{MAPF solving via SAT compilation (SMT-CBS).} \label{alg-SATCompilation}
\end{footnotesize}
\end{algorithm}

\section{Beyond Simple Time Expansion}

MIP as the target formalism allows for reasoning about both integer and real-valued decision variables which opens opportunities to use decision variables with completely different meaning than in CSP and SAT. Bounding linear objectives is also easier in MIP than in SAT as MIT can natively work with numbers.

The model suggested in \cite{DBLP:conf/ijcai/LamBHS19} considers a large but finite pool of paths $\Pi(a_i)$ for each agent $a_i \in A$ connecting its start and goal vertex. Decision variables $\lambda_{i,\pi} \in [0,1]$ determine the proportion of path $\pi \in \Pi(a_i)$ being selected by the agent. Constraints ensure that agents use at least one path and the overall cost of path is minimized resulting in the following initial linear program $\mathit{LP}$:
\begin{equation}
\label{eq_MIP-min}
min \sum_{a_i \in A} {\sum_{\pi \in \Pi(a_i)} \lambda_{i,\pi} \mathit{cost}}(\pi)
\end{equation}

\begin{equation}
\label{eq_MIP-fea1}
{\sum_{\pi \in \Pi(a_i)} \lambda_{i,\pi}} \geq 1 \;\;\;\; \forall a_i \in A
\end{equation}

\begin{equation}
\label{eq_MIP-fea2}
\lambda_{i,\pi} \geq 0 \;\;\;\; \forall a_i \in A, \forall \pi \in \Pi(a_i)
\end{equation}

The program represents an incomplete encoding of the input instance as collisions between agents are not forbidden initially. Collision resolution is done lazily similarly as in the SAT-based approach. However the important difference here is that the pool of candidate paths is extended together with adding collision avoidance constraints.

Once a vertex collision in $v \in V$ is detected in current (fractional) solution $\{\lambda_{i,\pi}\}$, that is the following inequality holds for $v$:

\begin{equation}
\label{eq_MIP-collision1}
{\sum_{a_i \in A} \sum_{\pi \in \Pi(a_i)} s_v^\pi \lambda_{i,\pi}} > 1,
\end{equation}

where $s_v^\pi \in \{0,1\}$ indicates selection of vertex $v$ by path $\pi$ (1 means the vertex is selected), corresponding collision elimination constraint for vertex $v$ is included into the encoding; that is, the following constraint is added:

\begin{equation}
\label{eq_MIP-collision2}
{\sum_{a_i \in A} \sum_{\pi \in \Pi(a_i)} s_v^\pi \lambda_{i,\pi}} \leq 1,
\end{equation}

Now cost penalties are introduced for agents using vertex $v$ where the collision happened and the pool of paths is extended with at least one better path for each agent considering the paths' costs and the penalties. Compared to the SAT-based approach, the extension of the set of candidate paths is more general, more fine grained, and provides room to integrate heuristics that include more promising paths first. The SAT-based compilation increases the bound of the sum-of-costs (or other objective) instead, which allows considering more paths satisfying the bound in future iterations, but such extension adds many new paths all at once without distinguishing if some of them are more promising than others.

On the other hand, the MIP-based approach compared to SAT-based compilation requires to deal with fractional solutions, which adds non-trivial complexity to the high level solving process. As shown in the framework of the MIP-based approach, Algorithm \ref{alg-MIPCompilation}, elimination of fractionality of solutions requires to implement {\em branch-and-bound} search represented in the code by a non-deterministic choice (line 18). In contrast to this, all non-polynomial time effort is left to the SAT solver in the SAT-based approach.

Experiments show that the method performs well even in its vanilla variant. This can be attributed to the strength of the target MIP solver and to the design of decision variables that are particularly suitable for it.

\begin{algorithm}[t]
\begin{footnotesize}
\SetKwBlock{NRICL}{Lazy-MIP-MAPF($\mathcal{M}$)}{end} \NRICL{
    $\Pi \gets$ initialize-Paths-Pool($\mathcal{M}$)\\ 
    $\mathit{constraints} \gets \emptyset$\\
    \While {$\mathit{TRUE}$} {   
        $(\mathit{paths},\Pi',\mathit{constraints}) \gets$ \\ Solve-Using-Path-Pool($\mathcal{M}$, $\Pi$, $\mathit{constraints}$)\\ 
        \If {$\mathit{paths} \neq \mathit{UNSAT}$}{
            \Return $\mathit{paths}$\\
        }
        $\Pi \gets \Pi'$\\
     }
}

\SetKwBlock{NRICL}{Solve-Using-Path-Pool($\mathcal{M}$,$\Pi$,$\mathit{constraints}$)}{end} \NRICL{
	\While {$\mathit{TRUE}$} {
          $\mathit{LP} \gets$ encode-MIP($\mathcal{M}$, $\Pi$, $\mathit{constraints}$)\\
          ${\{\lambda_{i,\pi}\}}_{a_i \in A,\pi \in \Pi(a_a)} \gets$ consult-MIP-Solver($\mathit{LP}$)\\         
          $\mathit{conflicts} \gets$ validate($\mathcal{M}$,$\{\lambda_{i,\pi}\}$)\\
          
	    \If{$\mathit{conflicts} = \emptyset$}{
          	 \If{$\exists \pi \in \Pi, a_i \in A$ such that $\lambda_{i,\pi} > 0 \wedge \lambda_{i,\pi} < 1$}{
          	     $\mathit{resolvers} \gets$ resolve-Fractionality($\mathit{LP}$,$\{\lambda_{i,\pi}\}$)\\
          	     {\bf let}  $r \in \mathit{resolvers}$ /* non-deterministic */\\
          	     $\mathit{constraints} \gets \mathit{constraints} \cup$ $\{r\}$\\
          	     \Return $(\mathit{UNSAT}, \Pi, \mathit{constraints})$
          	 }
          	 \Else{
            	    $\mathit{paths} \gets$ interpret($\mathcal{M}$,${\{\lambda_{i,\pi}\}}$)\\
           	   \Return $(\mathit{paths}, \Pi, \mathit{constraints})$
           	 }	    
          } 
          \For{each $c \in \mathit{conflicts}$}{
              $\mathit{constraints} \gets$ \\ $\mathit{constraints} \cup$ eliminate-Conflict($c$)\\
          }
          \For{$i=1,2,...,k$}{
             $\Pi_{better} \gets$ find-Better-Paths($i$,$\{\lambda{_{i,\pi}}\}$,$\mathit{constraints}$)\\
             $\Pi(a_i) \gets \Pi(a_i) \cup \Pi_{better}$\\
          }
      }
}
\caption{MAPF solving via MIP compilation.} \label{alg-MIPCompilation}
\end{footnotesize}
\end{algorithm}

\section{SAT-based vs. MIP-based Compilation}

As the pseudo-codes of the MIP-based and SAT-based compilation schemes suggest there are different opportunities how to enhance each approach with MAPF specific heuristics and pruning techniques such as {\em symmetry breaking} \cite{DBLP:conf/aips/0001GHS0K20} or {\em mutex reasoning} \cite{DBLP:conf/aips/ZhangLSKK20}. In this regard, MIP-based scheme is more open for integration of domain specific improvements as more decisions are made at the high-level, namely branching strategy where heuristics can be included, paths pool refinement that could further guide the search can be modified at the high level. Altogether, the MIP solver is treated more like a white-box.

In contrast to this, SAT-based approach still leaves lot of decisions on general purpose SAT-solver into which it is difficult to include any MAPF specific heuristics without changing the implementation of the solver. On the other hand, to role of the solver is bigger as it solves without any intervention from the high-level entire NP-hard component of the problem. In the MIP-based approach, the MIP solver solves independently a linear problem which is done in polynomial time while the hard exponential-time part is solved in cooperation with the high-level.

Significant weak point in SAT-based compilation is arithmetic and consequently difficult handling of the MAPF objectives in the target formalism. Bounding the objective in the yes/no environment of SAT must be done via modeling arithmetic circuits inside the formula which has twofold difficulties. The circuit is often large in the number of variables we need to bound and Boolean constraint propagation is usually less efficient across the circuit. MIP on the other hand can express common objectives used in MAPF as its own objective in straightforward way.

\section{Conclusion}
We summarized and compared main ideas of recent compilation-based approaches to MAPF, the MIP-based and SAT-based solvers. We also describe and analyze important milestones advancing the field to the current state-of-the-art. Our summary highlights important common features of both approaches such as lazy conflict elimination, but also focuses on significant differences such as the need in SAT to deal with cost bounds at the level of Boolean formula or the need in MIP to eliminate fractional values of decision variables.

As it is clear from aspects we focus on, we attribute great importance to design choices of how the compilation algorithms operate (e.g. eager vs. lazy style). In our perspective, algorithmic aspects of compilation approach is has more significant impact than concrete encoding of rules MAPF in the target formalism. More important is the design of solving algorithm that uses the encoding as its building block.

\subsection{Future Prospects}

The concept of laziness makes compilation flexible so it is applicable in various generalized variants of MAPF such as MAPF with continuous time \cite{DBLP:conf/ijcai/AndreychukYAS19} as shown for SAT \cite{DBLP:conf/icaart/Surynek20}. Adapting the MIP approach for MAPF with continuous time is likely possible too.

Another room for future improvements comes from the fact that the solver both in the SAT-based and MIP-based compilation runs uninterrupted until the solution to the encoded instance is found or it is proved that there is no solution. This could be inefficient in case of the incomplete encodings since there could be a conflict induced by partial assignment of decision variables. Checking partial assignments for conflicts could be the next step in compilation-based approaches. The framework for SAT is already known as DPLL(T) \cite{DBLP:conf/fmcad/KatzBTRH16}.


\section*{Acknowledgments}
This research has been supported by GA\v{C}R - the Czech Science Foundation, grant registration number 19-17966S.

\bibliographystyle{named}
\bibliography{references}

\end{document}